\documentclass[sigconf,nonacm,natbib=false]{acmart}
\usepackage[utf8]{inputenc}
\usepackage[T1]{fontenc}
\usepackage[USenglish]{babel}
\usepackage{graphicx}
\usepackage{amsmath,amsfonts}
\usepackage{csquotes}
\usepackage[
  backend=biber,
  style=authoryear,
  sorting=nty,
  natbib=true,
  citetracker=true,
  maxcitenames=5,
  mincitenames=1,
  maxbibnames=99
]{biblatex}
\usepackage{booktabs}
\usepackage{multirow}
\usepackage{algorithm}
\usepackage{algpseudocode}

\AtEveryCitekey{%
  \ifciteseen
    {\ifnumgreater{\value{labelname}}{2}
       {\defcounter{maxnames}{1}}
       {}}
    {}}

\graphicspath{{./images/}}

\def\projecttitle{DGLight: DQN-Guided GRPO Fine-Tuning of Large Language Models for Traffic Signal Control}
\title{\projecttitle}
\author{Chenbo Yu}
\affiliation{%
  \institution{National University of Singapore}
  \country{Singapore}
}
\email{yu.chenbo@u.nus.edu}

\begin{document}
\begin{abstract}
Traffic signal control (TSC) plays a central role in reducing congestion and maintaining urban mobility. This dissertation introduces \textit{DGLight}, a critic-guided reinforcement-learning framework for adapting a pretrained large language model to TSC. \textit{DGLight} first trains a \textit{CoLight}-based Deep Q-Network critic to estimate traffic-aware action values from structured intersection states, then uses the frozen critic to score candidate language-model actions and optimize the policy with Group Relative Policy Optimization (GRPO). The resulting controller maps traffic states to interpretable reasoning traces and signal decisions while learning from dense per-state supervision rather than raw cumulative environment rewards. Experiments on TSC benchmarks covering Jinan and Hangzhou show that \textit{DGLight} is the strongest overall method among the compared LLM-based controllers, remains competitive with strong RL baselines, and transfers well to city datasets not used to fit the critic. Qualitative examples further show that the model's generated reasoning is interpretable and aligned with the chosen signal phase. The project code is available at \url{https://github.com/yyccbb/FYP_LLMTSC}.
\end{abstract}


\maketitle

\section{Introduction}
Rapid urbanization and accelerating motorization have placed unprecedented strain on urban transportation systems worldwide. As cities densify, limited road infrastructure must accommodate ever-growing travel demand, leading to persistent congestion, unreliable travel times, and increased environmental externalities. Inefficient traffic flow not only wastes fuel and economic productivity but also amplifies air pollution and degrades quality of life, particularly in densely populated urban corridors. \par
Traffic signal control (TSC) aims to coordinate phase sequences and durations at signalized intersections to maximize throughput and reduce congestion. Classical fixed-time methods use the Webster formula \parencite{webster} to preset signal timings based on historical traffic data collected from the traffic network at different times. In contrast, the max-pressure control framework \parencite{MERCADER2020275} computes a pressure value for each signal phase from the difference between queue lengths on opposing approaches and selects the phase with the highest pressure; this strategy can be shown to maximize network throughput.\par
Reinforcement-learning (RL) methods model TSC as a Markov decision process in which an agent observes traffic states (for example, lane queue lengths and delays), chooses an action (phase switch) and receives a reward linked to throughput or delay reduction. Deep RL further parameterizes the control policy using neural networks, enabling the agent to approximate complex, high-dimensional decision functions. Such methods have demonstrated strong empirical performance across a wide range of simulated traffic scenarios. However, they also introduce notable challenges. In particular, learned policies often exhibit limited generalization, as models trained in one traffic environment exhibit degraded performance when deployed in different cities or under unseen conditions. Furthermore, the reliance on opaque neural representations reduces interpretability, making it difficult to explain or verify decision-making processes. This lack of transparency poses a barrier to real-world adoption, where reliability and accountability are critical.\par
Large Language Models (LLMs) represent a new paradigm in artificial intelligence, combining strong reasoning ability with the capacity to generate interpretable, text-based outputs. These properties make LLMs promising candidates for traffic signal control, where decision transparency and adaptability are highly desirable. By framing control as a reasoning process, LLM-based agents can potentially provide both actionable policies and interpretable explanations, bridging the gap between performance and interpretability. However, this potential is not yet fully realized in practice. Achieving real-time responsiveness typically requires deploying smaller-scale models or mixture-of-experts (MoE) architectures with compact expert networks. Without sufficient domain adaptation, such models often fail to capture complex traffic dynamics, leading to suboptimal decisions.\par
To this end, I introduce \textit{DGLight}, a critic-guided reinforcement-learning framework for adapting a pretrained large language model to traffic signal control. The framework combines a Deep Q-Network (DQN) critic with GRPO-based policy optimization so that textual traffic prompts can be scored using dense, traffic-aware action values. In this way, \textit{DGLight} trains the language model to produce both an interpretable reasoning trace and a valid signal decision, transferring value-based traffic knowledge into the policy to improve decision quality across traffic settings.\par
The main contributions of this dissertation are:
\begin{itemize}
    \item First, I propose the \textit{DGLight} framework, which unifies a value-based traffic critic and LLM fine-tuning within a single critic-guided training pipeline. 
    \item Second, I trained a TSC controller with \textit{DGLight} by adapting a \textit{Llama3-8b} traffic-signal model through the proposed pipeline, yielding an interpretable language-based policy for multi-intersection control. 
    \item Third, I conduct comprehensive experiments on standard TSC benchmarks, including baseline comparisons, transfer evaluation on unseen city datasets, interpretability analysis, and ablation studies, to validate the effectiveness of the proposed method.
\end{itemize} 

\section{Related Work}
\textbf{Traffic Signal Control.} Early TSC methods such as \textit{FixedTime} \parencite{webster} rely on predefined cycle lengths and manually designed phase splits for each intersection. Although these approaches are simple and stable in practice, they adapt poorly to time-varying traffic demand. \textit{MaxPressure} \parencite{MERCADER2020275} improves adaptability by selecting phases according to the queue imbalance between incoming and outgoing directions, making it more responsive than fixed plans. However, both \textit{FixedTime} and \textit{MaxPressure} remain constrained by handcrafted rules or preset logic, which limits their flexibility under complex and highly dynamic traffic scenarios.

Recent progress in reinforcement learning has led to a series of RL-based TSC methods. \textit{CoLight} \parencite{10.1145/3357384.3357902} models network-level cooperation among intersections through graph attention, enabling each signal to coordinate with surrounding intersections instead of acting independently. \textit{CoSLight} \parencite{ruan2024coslightcooptimizingcollaboratorselection} further extends this idea by jointly optimizing signal control and collaborator selection, so that each intersection can dynamically decide which neighboring intersections are most useful for coordination under current traffic conditions. \textit{MPLight} \parencite{Chen_Wei_Xu_Zheng_Yang_Xiong_Xu_Li_2020} is a decentralized and pressure-aware RL approach designed for large-scale road networks, using parameter sharing to improve scalability and training efficiency. Although these methods achieve strong performance in per city simulation, RL-based controllers face challenges in generalization and interpretability, hence deployment robustness in real-world environments.

To address these limitations, recent studies have begun exploring the integration of large language models (LLMs) into TSC. \textit{LLM-Assisted Light} \parencite{wang2024llmassistedlightleveraging} places an LLM at the center of control and augments it with perception and decision-support tools so that it can combine live traffic observations with established TSC methods and remain robust under emergency, roadblock, and sensor-outage scenarios. \textit{iLLM-TSC} \parencite{pang2024illmtscintegrationreinforcement} instead adopts a hybrid design in which an RL controller first proposes an action and an LLM then verifies or adjusts that action when observations are degraded or rare events such as packet loss and emergency vehicles fall outside the original RL formulation. \textit{LATS} \parencite{zhang2026latslargelanguagemodelassisted} further explores a teacher-student setting in which an LLM teacher provides rich semantic features during multi-agent RL training, while a smaller student controller learns to emulate those features and is deployed without the LLM at inference time. \textit{LLMLight} \parencite{lai2024llmlightlargelanguagemodels} moves closer to end-to-end LLM control by adapting its LightGPT backbone through imitation fine-tuning and critic-guided policy refinement so that the model can make direct signal decisions with improved interpretability and cross-scenario performance. \textit{Traffic-R1} \parencite{zou2025trafficr1reinforcedllmsbring} extends this trend with a 3B RL-finetuned foundation model that combines expert-aligned offline training, online self-exploration, and asynchronous multi-agent communication to achieve strong zero-shot and out-of-distribution TSC performance. Compared with end-to-end LLM finetuning, LLM+RL hybrids may remain inferior because they still inherit the ceiling and failure modes of the underlying RL controller. Existing end-to-end finetuning methods also have notable limitations: In \textit{LLMLight}, the critic-guided policy refinement objective is weaker than modern policy-gradient methods because the reward/advantage magnitude does not directly participate in loss computation. \textit{Traffic-R1}, meanwhile, relies on substantial human effort to label the offline finetuning data, and its fixed-horizon cumulative queue-based reward assigns the same coarse mean reward to all steps within the horizon.

\textbf{RL finetuning of LLMs.} RL-based preference alignment \parencite{ouyang2022traininglanguagemodels} adapts a pretrained LLM using scalar or pairwise feedback so that the policy is optimized toward preferred responses rather than merely imitating target outputs. Among modern policy optimization methods, TRPO \parencite{schulman2015trustregionpolicyoptimization} constrains updates within a trust region to stabilize improvement, PPO \parencite{schulman2017proximalpolicyoptimizationalgorithms} replaces this with a clipped surrogate objective that is simpler and more practical at scale, and GRPO \parencite{shao2024deepseekmath} further removes the separate value model by computing group-relative advantages over sampled responses, which makes it especially attractive for LLM training. Closely related to these RL-style methods, DPO \parencite{rafailov2023directpreferenceoptimization} performs direct preference optimization on chosen--rejected response pairs and avoids explicit reward-model fitting and online rollout collection. This line of work is particularly important because preference-aligned RL can train reasoning ability rather than only answer imitation: through exploration, comparison, and preference-based updates, the model learns to assign higher probability to response trajectories that exhibit stronger multi-step reasoning while remaining close to a reference policy through KL regularization, as emphasized in RLHF-style alignment and recent reasoning-oriented RL studies \parencite{ouyang2022traininglanguagemodels,shao2024deepseekmath,guo2025deepseekr1}. Recent systems such as \textit{DeepSeek-R1} \parencite{guo2025deepseekr1} show that large-scale self-exploration can substantially improve reasoning behavior, making RL preference alignment increasingly relevant for domain-specific agents that must optimize long-horizon objectives.

\section{Preliminaries}

\textsc{Definition 1.} \textbf{\textit{Road Network.}} A road network is represented as a directed graph consisting of intersections $\mathcal{V}$ and lanes $\mathcal{L}$. The lane set is grouped into three categories: go-through lanes ($\mathcal{L}_{\mathrm{go}}$), left-turn lanes ($\mathcal{L}_{\mathrm{left}}$), and right-turn lanes ($\mathcal{L}_{\mathrm{right}}$). Each lane connects adjacent intersections and is partitioned into segments $S = \{s_1, \ldots, s_n\}$ according to its distance from the intersection.

\textsc{Definition 2.} \textbf{\textit{Signal Phase.}} At each signal-switching step, the controller of an intersection chooses a signal phase from a predefined action set $A = \{a_1, \ldots, a_m\}$. A phase is written as $a = \mathrm{set}(\mathcal{L}_{\mathrm{allow}})$, where $\mathcal{L}_{\mathrm{allow}}$ denotes the collection of lanes that are allowed to move simultaneously without conflict, meaning these lanes receive green signals while conflicting lanes remain red.

\textsc{Definition 3.} \textbf{\textit{Traffic Signal Control System.}} The traffic signal control system consists of multiple agents $\Pi = \{\pi_1, \ldots, \pi_n\}$, with each agent assigned to one of the $n$ intersections in the road network. For intersection $i$, agent $\pi_i$ makes its signal control decision by observing the number of vehicles in each segment of the lanes approaching that intersection, thereby forming a segment-level view of the local traffic state.

\textsc{Problem Statement.} \textbf{\textit{LLM Reasoning for Traffic Signal Control.}} For each intersection $i$ in the road network, an LLM serves as agent $\pi_i$. At decision step $t$, the agent takes the observation $O_{i,t}$, the task description $d_{\mathrm{task}}$, and the action space $A$ as input, and outputs both reasoning content and a signal phase decision. The TSC problem is therefore formulated as selecting a phase $a_{i,t} \in A$ while generating reasoning output $R_{i,t}$ under the given observation and task specification. The system dynamics are defined as:

\begin{equation}
(R_{i,t}, a_{i,t}) = \pi_{\theta_i}(O_{i,t}, d_{\mathrm{task}}, A)
\end{equation}

where $O_{i,t}$ is the observation, $d_{\mathrm{task}}$ is the task description, $A$ is the action space, $R_{i,t}$ is the reasoning output, and $a_{i,t}$ is the selected phase.

\section{Methodology}
\begin{figure*}[t]
    \centering
    \includegraphics[width=1\linewidth]{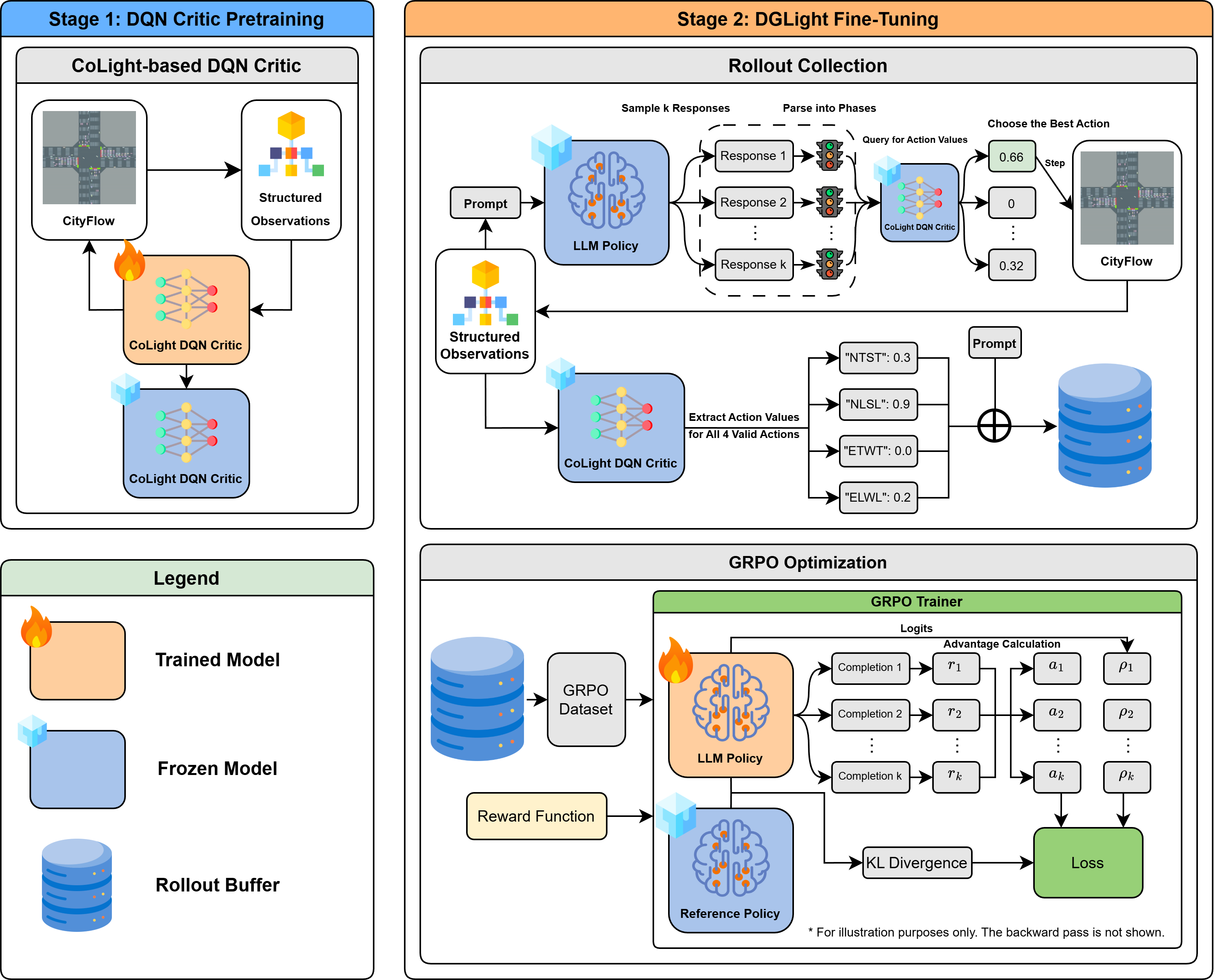}
    \caption{Overview of the two-stage \textit{DGLight} pipeline.}
    \label{fig:dglight_pipeline}
\end{figure*}

This section introduces \textit{DGLight}, a two-stage reinforcement-learning pipeline for adapting a pretrained large language model to traffic signal control. In the first stage, a CoLight-based DQN critic is trained to estimate action values for signal-phase decisions when provided with structured environment states. In the second stage, the frozen critic scores candidate LLM responses during rollout and the scores are used as the reward signal for GRPO-based policy optimization. Refer to Figure~\ref{fig:dglight_pipeline} for the pipeline architecture.

\subsection{Pretraining the CoLight-Based DQN Critic}
The critic used in \textit{DGLight} is based on \textit{CoLight} \parencite{10.1145/3357384.3357902}, a graph-attentive deep Q-learning model for network-level traffic signal control. \textit{CoLight} is employed as a pretrained value estimator so that the subsequent GRPO stage can query a traffic-aware action-value function rather than rely exclusively on the raw environment reward. Given the current environment state, the critic predicts action values for each intersection and thereby provides a dense estimate of the quality of candidate signal phases under the prevailing traffic situation.

The \textit{CoLight} critic is composed of three main components. The first component is an observation embedding module that transforms the local traffic state of each intersection into a latent feature representation. The second component consists of one or more multi-head graph-attention layers, which propagate information across neighboring intersections and produce attention-weighted network representations. The final component is a Q-value prediction layer that maps the aggregated features to one scalar value for each admissible signal phase. This architecture combines local state encoding, inter-intersection coordination, and value-based action estimation within a single model.



Training follows the standard value-based reinforcement learning procedure. The model interacts with the traffic environment to collect state transitions, actions, rewards, and subsequent states ($(s,a,r,s')$), where the stepwise reward is defined to be inversely proportional to queue lengths at each intersection. These transitions are stored in replay memory for off-policy updates. The optimization objective minimizes the Bellman error
\[
\mathcal{L}_{\mathrm{Bellman}}(\theta)
=
\mathbb{E}_{(s,a,r,s')\sim \mathcal{D}}
\left[
\left(
r + \gamma \max_{a'} Q_{\theta^-}(s',a') - Q_{\theta}(s,a)
\right)^2
\right],
\]
where $\mathcal{D}$ denotes the replay buffer, $\theta$ denotes the online network parameters, and $\theta^-$ denotes the target network parameters. Experience replay improves sample efficiency, and the target network improves optimization stability.

\textbf{Disclaimer:} The critic model, its training objective, and its overall learning logic are not original contributions of this dissertation; they are inherited from prior \textit{CoLight} research. My contribution to this component is limited to re-implementing the model in PyTorch for this codebase, whereas the original \textit{CoLight} implementation was based on TensorFlow.

\subsection{LLM Prompting and Action Sampling}
At each signal-switching step $t$, each intersection $i$ is converted into a textual prompt $x_{i,t}$ that summarizes the current traffic condition together with the task instruction and the available signal phases. The prompt is constructed from the environment state used by the controller including the current phase and queue information. Task description is formatted so that the model produces both human-readable reasoning and a final phase choice. A sample prompt is provided in Appendix~\ref{appendix:sample_prompt}.

Given prompt $x_{i,t}$, the policy $\pi_\theta$ samples a group of $k$ candidate responses
\[
\{y_{i,t}^{(1)}, y_{i,t}^{(2)}, \dots, y_{i,t}^{(k)}\}.
\]
Each response is then mapped to a signal phase through a deterministic parser:
\[
a_{i,t}^{(j)} = \operatorname{Parse}(y_{i,t}^{(j)}), \qquad j = 1, \dots, k.
\]
If the response fails to specify a valid action in the predefined action space, it is treated as an invalid output and assigned a fallback reward during training. This allows the method to preserve free-form reasoning generation while still enforcing a discrete control interface at execution time.

\subsection{DQN-Guided Action Scoring in Rollout Collection}
The central idea of \textit{DGLight} is to replace discounted raw environment rewards with action value estimates from the pretrained DQN critic.

\textbf{Stepwise action value assignment.} For the current traffic state at intersection $i$, the frozen critic produces a vector of estimated returns over all admissible signal phases:
\[
\mathbf{q}_{i,t} = \bigl[ Q_\phi(s_{i,t}, a) \bigr]_{a \in A},
\]
where $s_{i,t}$ is the structured traffic state and $Q_\phi$ is the pretrained DQN critic. The critic therefore acts as a traffic-aware evaluator that maps the current state to a vector of action values that reflect both spatial and temporal traffic context.

Each sampled LLM response is scored by the critic through its parsed action:
\[
r_{i,t}^{(j)} =
\begin{cases}
Q_\phi\!\left(s_{i,t}, a_{i,t}^{(j)}\right), & \text{if } a_{i,t}^{(j)} \in A, \\
r_{\mathrm{invalid}}, & \text{otherwise},
\end{cases}
\]
where $r_{\mathrm{invalid}}$ is a fixed fallback reward for malformed or out-of-range decisions. 

\textbf{Environment stepping.} Among the sampled candidates, the action actually executed in the environment is the highest-scoring sampled action,
\[
j_{i,t}^{*} = \arg\max_{j \in \{1,\dots,k\}} r_{i,t}^{(j)},
\qquad
\hat a_{i,t} = a_{i,t}^{(j_{i,t}^{*})}.
\]
Thus, the environment rollout is still driven by sampled LLM outputs, but execution is guided by the critic's action-value estimate.

This design has a major advantage that the critic-derived reward incorporates richer spatial and temporal information than short-horizon cumulative rewards computed directly from the environment. Because the CoLight-based critic aggregates information from neighboring intersections through graph attention and is trained with Bellman backups under a discount factor, $Q_\phi(s_{i,t}, a)$ reflects both network interactions and the expected downstream consequences of choosing action $a$. 

\subsection{GRPO-Based Policy Optimization}
After rollout collection, each intersection-step pair contributes a training record consisting of the prompt and the full critic score vector,
\[
\mathcal{D}_{i,t} = \bigl(x_{i,t}, \mathbf{q}_{i,t}\bigr).
\]
Importantly, the stored supervision is not just the score of the executed action. Instead, the entire per-action vector is retained so that any newly sampled response for the same prompt can be rewarded according to the action it proposes. During GRPO training, a completion $y$ sampled for prompt $x_{i,t}$ is assigned reward
\[
R\!\left(x_{i,t}, y; \mathbf{q}_{i,t}\right) =
\begin{cases}
\mathbf{q}_{i,t}[a(y)], & \text{if } a(y) \in A, \\
r_{\mathrm{invalid}}, & \text{otherwise},
\end{cases}
\]
where $a(y)$ denotes the parsed action of completion $y$. The resulting reward values are then normalized within each sampled group, producing relative advantages that drive the GRPO update.

This optimizer is particularly well suited to the present setting because it provides a natural way to train an LLM using reward signals from an already established value-based traffic-control critic. For a single traffic prompt, multiple completions can be sampled efficiently, parsed into candidate signal phases, and scored by the frozen DQN critic, after which GRPO uses the relative critic rewards within the sampled group to improve the policy. This way, critic-guided GRPO transfers decision-making knowledge from the value-based controller into the LLM. Although the critic is trained on traffic data from a specific city, the LLM is not fine-tuned to overfit any single city. Rather, it is trained to have stronger reasoning patterns for mapping textual traffic states to better signal actions. This strengthens the model's traffic decision-making capacity while preserving transferability.

\subsection{End-to-End DGLight Training Procedure}
Algorithm~\ref{alg:dglight_training} summarizes the complete \textit{DGLight} training loop.

\begin{algorithm}[h]
    \caption{End-to-end \textit{DGLight} training}
    \label{alg:dglight_training}
    \begin{algorithmic}[1]
        \Require traffic environment $\mathcal{E}$, pretrained LLM policy $\pi_\theta$, critic architecture $Q_\phi$, candidate count $k$, invalid-action reward $r_{\mathrm{invalid}}$
        \State Train the CoLight-based DQN critic $Q_\phi$ on traffic transitions
        \State Freeze critic parameters $\phi$
        \For{each policy-training episode}
            \State Reset environment $\mathcal{E}$
            \State Initialize empty GRPO record set $\mathcal{D}_{\mathrm{GRPO}}$
            \While{the rollout episode is not finished}
                \State Initialize empty executed action set $\hat{\mathbf{a}}_t$
                \For{each active intersection $i$}
                    \State Construct textual prompt $x_{i,t}$ from the current traffic state $s_{i,t}$
                    \State Sample candidate responses $\{y_{i,t}^{(1)}, \dots, y_{i,t}^{(k)}\} \sim \pi_\theta(\cdot \mid x_{i,t})$
                    \State Parse each response into a signal phase $a_{i,t}^{(j)} \gets \operatorname{Parse}(y_{i,t}^{(j)})$
                    \State Query the frozen critic to obtain per-action values $\mathbf{q}_{i,t} \gets [Q_\phi(s_{i,t}, a)]_{a \in A}$
                    \For{each sampled response $j \in \{1,\dots,k\}$}
                        \If{$a_{i,t}^{(j)} \in A$}
                            \State $r_{i,t}^{(j)} \gets Q_\phi(s_{i,t}, a_{i,t}^{(j)})$
                        \Else
                            \State $r_{i,t}^{(j)} \gets r_{\mathrm{invalid}}$
                        \EndIf
                    \EndFor
                    \State $j_{i,t}^{*} \gets \arg\max_j r_{i,t}^{(j)}$
                    \State $\hat{a}_{i,t} \gets a_{i,t}^{(j_{i,t}^{*})}$ and add $\hat{a}_{i,t}$ to $\hat{\mathbf{a}}_t$
                    \State Store $(x_{i,t}, \mathbf{q}_{i,t})$ in $\mathcal{D}_{\mathrm{GRPO}}$
                \EndFor
                \State Step the environment with joint action set $\hat{\mathbf{a}}_t$
            \EndWhile
            \State Fine-tune policy $\pi_\theta$ on $\mathcal{D}_{\mathrm{GRPO}}$ with GRPO
        \EndFor
        \State \Return adapted policy $\pi_\theta$
    \end{algorithmic}
\end{algorithm}

In practice, the rollout collection and GRPO optimization stages are decoupled. A full rollout episode is first collected with the frozen critic, and the resulting prompt-score records are then used as a separate training set for the subsequent GRPO update. This separation enables finer-grained optimization, since multiple manual updates can be performed on the same episodic rollout data.

\subsection{Comparison Variant}
For completeness, I also implemented a \textit{Joint-Scored GRPO (JS-GRPO)} variant as a comparison method. Instead of using direct critic $Q$-values to score sampled actions, that variant estimates action quality indirectly from short-horizon joint rollout returns over snapshot-forked simulations, and then projects those returns back to per-intersection action scores for GRPO training. Because this variant is included only to test whether indirect joint scoring can outperform direct critic guidance, I do not treat it as a co-equal methodological contribution in the main text. Its detailed design is provided separately in Appendix~\ref{appendix:joint_scored_grpo}.

\section{Experiments}
In this section, I evaluate \textit{DGLight} to address the following research questions:
\begin{itemize}
    \item \textbf{RQ1: (Effectiveness)} Can \textit{DGLight} outperform traditional, RL-based, and existing LLM-based traffic signal control methods on standard TSC datasets?
    \item \textbf{RQ2: (Transferability)} Does \textit{DGLight} retain strong performance when evaluated on city datasets that were not used to train the DQN critic?
    \item \textbf{RQ3: (Interpretability)} Are the decisions produced by \textit{DGLight} interpretable, in the sense that its generated reasoning reflects the traffic conditions that motivate the selected phase?
    \item \textbf{RQ4: (Ablation Study)} How much improvement comes from critic-guided GRPO, as compared with baseline and intermediate models?
\end{itemize}

\subsection{Experimental Settings}
This subsection describes the datasets, simulation environment, baselines, evaluation metrics, and training configuration used throughout the experiments.

\begin{table}[h]
    \centering
    \caption{Statistics of traffic flow datasets.}
    \label{tab:dataset_summary}
    \resizebox{1\linewidth}{!}{
    \begin{tabular}{ccccccc}
        \hline
        \multirow{2}{*}{Dataset} & \multirow{2}{*}{Structure} & \multirow{2}{*}{Vehicles} & \multicolumn{4}{c}{Arrival rate (vehicles/5min)} \\ \cline{4-7} 
        &  &  & \multicolumn{1}{l}{Mean} & \multicolumn{1}{l}{Std} & \multicolumn{1}{l}{Max} & \multicolumn{1}{l}{Min} \\ \hline
        Jinan1 & \multirow{3}{*}{$3 \times 4$} & 6295 & 523.67 & 98.52 & 671 & 255 \\
        Jinan2 &  & 4365 & 362.83 & 74.81 & 493 & 236 \\
        Jinan3 &  & 5494 & 456.92 & 160.87 & 569 & 362 \\ \hline
        Hangzhou1 & \multirow{2}{*}{$4 \times 4$} & 2983 & 247.68 & 40.44 & 332 & 211 \\
        Hangzhou2 &  & 6984 & 581.08 & 318.43 & 1145 & 202 \\ \hline
    \end{tabular}
    }
\end{table}

\subsubsection{Dataset}
Experiments are conducted on standard traffic signal control benchmarks. Table~\ref{tab:dataset_summary} shows the final dataset statistics.

\subsubsection{Environment}
All methods are evaluated in an environment built on top of \textit{CityFlow} \parencite{tang2019cityflowcityscalebenchmarkmultitarget}, a scalable microscopic traffic simulator widely used in traffic signal control research. For each dataset, the simulator loads the corresponding road-network and traffic-flow configuration and exposes lane-level observations at every control step. In my experiments, each intersection connects four lane sections: east, west, north, and south. The control action space therefore contains four signal phases: \textit{ETWT} (through movements from east and west), \textit{ELWL} (left turns from east and west), \textit{NTST} (through movements from north and south), and \textit{NLSL} (left turns from north and south). Right-turn movements are always allowed in the simulation. Each controller selects one signal phase for every controlled intersection, after which the environment advances under a shared signal-switching interval. The green phase duration is fixed at 30 seconds. Each green phase is followed by a 3-second yellow phase and a 2-second all-red interval to ensure safe signal transitions. Unless otherwise stated, all compared methods use the same simulator configuration, decision interval, and evaluation protocol so that performance differences can be attributed to the controller rather than the environment setup.

\subsubsection{Baselines}
The comparison includes traditional controllers, RL-based methods, and LLM-based methods.
\begin{itemize}
    \item \textbf{Traditional methods.} \textit{FixedTime} \parencite{webster} and \textit{MaxPressure} \parencite{MERCADER2020275}.
    \item \textbf{RL methods.} \textit{MPLight} \parencite{Chen_Wei_Xu_Zheng_Yang_Xiong_Xu_Li_2020}, \textit{AttendLight}, \textit{PressLight}, \textit{CoLight} \parencite{10.1145/3357384.3357902}, \textit{Efficient-CoLight}, and \textit{Advanced-CoLight}.
    \item \textbf{LLM-based methods.} \textit{LightGPT (Llama3-8b)} \parencite{lai2024llmlightlargelanguagemodels}, \textit{JS-GRPO-h3 (Llama3-8b)}, \textit{JS-GRPO-h6 (Llama3-8b)}, and the proposed \textit{DGLight (Llama3-8b)}. The two \textit{JS-GRPO} variants correspond to the comparison method described in Appendix~\ref{appendix:joint_scored_grpo}, instantiated with horizon lengths of 3 and 6, respectively.
\end{itemize}

\subsubsection{Metrics}
Following previous studies \parencite{pmlr-v162-zhang22ah}, the following metrics are used to evaluate the TSC agents trained with different methods:
\begin{itemize}
    \item \textbf{Average traveling time (ATT)} quantifies the average duration of all vehicles traveling from their origins to their respective destinations.
    \item \textbf{Average queue length (AQL)} is defined as the average number of queuing vehicles waiting in the road network.
    \item \textbf{Average waiting time (AWT)} quantifies the average queuing time of vehicles at every intersection in the road network.
\end{itemize}
Lower values indicate better traffic efficiency for all three metrics.

\subsubsection{Training Configuration}
All LLM-based methods use \textit{Llama3-8b} as the backbone model. For the proposed method, the initial checkpoint is inherited from the \textit{LLMLight} pipeline \parencite{lai2024llmlightlargelanguagemodels} rather than obtained through an additional supervised fine-tuning stage, since training a dedicated supervised model on high-quality traffic reasoning traces is beyond the computational budget of this dissertation. Starting from the \textit{LLMLight} base model isolates the contribution of the proposed critic-guided GRPO procedure while preserving a fair comparison with prior LLM-based controllers. The overall training framework then combines a pretrained CoLight critic with a GRPO-based policy optimization stage. The critic retains the uniform RL hyperparameters reported by \textcite{pmlr-v162-zhang22ah}, while the policy stage uses LoRA adaptation and distributed training. Table~\ref{tab:critic_training_config} and Table~\ref{tab:grpo_training_config} summarize the corresponding framework settings.

\begin{table}[t]
    \centering
    \caption{CoLight critic configuration.}
    \label{tab:critic_training_config}
    \small
    \begin{tabular}{p{0.42\linewidth}p{0.5\linewidth}}
        \toprule
        Component & Configuration \\
        \midrule
        Training dataset & Hangzhou-1 \\
        Optimizer & Adam with learning rate $1 \times 10^{-3}$ \\
        Discount factor $\gamma$ & 0.8 \\
        Replay buffer size & 12000 \\
        Sample size per round & 3000 \\
        Batch size & 20 \\
        Gradient epochs per round & 100 \\
        Target network update interval & every 5 rounds \\
        Exploration schedule & $\epsilon = \max(0.8 \cdot 0.95^r, 0.2)$ \\
        Reward & $-0.25 \,\cdot\,$queue length \\
        \bottomrule
    \end{tabular}
\end{table}

\begin{table}[t]
    \centering
    \caption{GRPO policy optimization configuration.}
    \label{tab:grpo_training_config}
    \small
    \begin{tabular}{p{0.42\linewidth}p{0.5\linewidth}}
        \toprule
        Component & Configuration \\
        \midrule
        Policy backbone & Llama3-8b \\
        LoRA configuration & rank = $8$, $\alpha = 16$, dropout $= 0.05$, bias = none \\
        Overall episodes & 2 \\
        \midrule
        Rollout environment & 3600-second episodes with a 30-second action interval \\
        Rollout group size $k$ & 4 \\
        Prompt / completion length & 2048 / 1024 tokens \\
        Rollout sampling & temperature $= 1.0$, top-$p = 1.0$, top-$k = 50$ \\
        \midrule
        GRPO optimizer and learning rate & default TRL optimizer, learning rate $= 1.0 \times 10^{-6}$ \\
        GRPO batch size / accumulation / epochs & 8 per device / 1 / 1 \\
        Invalid-action reward & 0.0 \\
        Distributed training setup & 2 H100-96GB GPUs, 16 CPUs, 128 GB RAM, 12-hour wall time \\
        \bottomrule
    \end{tabular}
\end{table}

\subsection{Results on TSC Datasets (RQ1)}
\begin{table*}[t]
    \centering
    \caption{Performance of traditional and RL-based baselines on the TSC datasets. Lower values are better. Best and second-best values are highlighted in bold and underlined, respectively.}
    \label{tab:tsc_baselines}
    \resizebox{\textwidth}{!}{%
    \begin{tabular}{l|ccc|ccc|ccc|ccc|ccc}
        \toprule
        & \multicolumn{9}{c}{Jinan} & \multicolumn{6}{c}{Hangzhou} \\
        \cmidrule(lr){2-10} \cmidrule(lr){11-16}
        Models & \multicolumn{3}{c}{1} & \multicolumn{3}{c}{2} & \multicolumn{3}{c}{3} & \multicolumn{3}{c}{1} & \multicolumn{3}{c}{2} \\
        \cmidrule(lr){2-4} \cmidrule(lr){5-7} \cmidrule(lr){8-10} \cmidrule(lr){11-13} \cmidrule(lr){14-16}
        & ATT & AQL & AWT & ATT & AQL & AWT & ATT & AQL & AWT & ATT & AQL & AWT & ATT & AQL & AWT \\
        \midrule
        \multicolumn{16}{c}{\textbf{Transportation Methods}} \\
        \midrule
        Random & 597.62 & 687.35 & 99.46 & 555.23 & 428.38 & 100.40 & 552.74 & 529.63 & 99.33 & 621.14 & 295.81 & 96.06 & 504.28 & 432.92 & 92.61 \\
        FixedTime & 481.79 & 491.03 & 70.99 & 441.19 & 294.14 & 66.72 & 450.11 & 394.34 & 69.19 & 616.02 & 301.33 & 73.99 & 486.72 & 425.15 & 72.80 \\
        MaxPressure & 281.58 & 170.71 & \underline{44.53} & 273.20 & 106.58 & \textbf{38.25} & 265.75 & 133.90 & \underline{40.20} & 325.33 & 68.99 & \underline{49.60} & 347.74 & 215.53 & \underline{70.58} \\
        \midrule
        \multicolumn{16}{c}{\textbf{RL Methods}} \\
        \midrule
        MPLight & 307.82 & 215.93 & 97.88 & 304.51 & 142.25 & 90.91 & 291.79 & 171.70 & 89.93 & 345.60 & 84.70 & 81.97 & 358.56 & 237.17 & 100.16 \\
        AttendLight & 291.29 & 186.25 & 61.73 & 280.94 & 115.52 & 52.46 & 273.02 & 144.05 & 55.93 & 322.94 & 66.96 & 55.19 & 358.81 & 239.05 & 72.88 \\
        PressLight & 291.57 & 185.46 & 50.53 & 281.46 & 115.99 & 47.27 & 275.85 & 148.18 & 54.81 & 364.13 & 98.67 & 90.33 & 417.01 & 349.25 & 150.46 \\
        CoLight & 279.60 & 168.53 & 58.87 & 274.77 & 108.28 & 54.14 & 266.39 & 135.08 & 53.33 & 322.85 & 66.94 & 61.82 & 342.90 & 212.09 & 99.74 \\
        Efficient-CoLight & \underline{277.11} & \underline{163.60} & \textbf{43.41} & \underline{269.24} & \underline{102.98} & \underline{39.74} & \underline{262.25} & \underline{129.72} & \textbf{39.99} & \underline{311.96} & \underline{58.20} & \textbf{36.83} & \underline{333.27} & \underline{189.65} & \textbf{61.70} \\
        Advanced-CoLight & \textbf{274.67} & \textbf{160.85} & 49.30 & \textbf{268.25} & \textbf{102.12} & 41.11 & \textbf{260.66} & \textbf{127.83} & 43.54 & \textbf{304.47} & \textbf{52.94} & \underline{41.75} & \textbf{329.16} & \textbf{186.34} & 76.59 \\
        \bottomrule
    \end{tabular}%
    }
\end{table*}

\begin{table*}[t]
    \centering
    \caption{Performance of LLM-based methods on the TSC datasets. Lower values are better. Best and second-best values are highlighted in bold and underlined, respectively.}
    \label{tab:tsc_llm_results}
    \resizebox{\textwidth}{!}{%
    \begin{tabular}{l|ccc|ccc|ccc|ccc|ccc}
        \toprule
        & \multicolumn{9}{c}{Jinan} & \multicolumn{6}{c}{Hangzhou} \\
        \cmidrule(lr){2-10} \cmidrule(lr){11-16}
        Models & \multicolumn{3}{c}{1} & \multicolumn{3}{c}{2} & \multicolumn{3}{c}{3} & \multicolumn{3}{c}{1} & \multicolumn{3}{c}{2} \\
        \cmidrule(lr){2-4} \cmidrule(lr){5-7} \cmidrule(lr){8-10} \cmidrule(lr){11-13} \cmidrule(lr){14-16}
        & ATT & AQL & AWT & ATT & AQL & AWT & ATT & AQL & AWT & ATT & AQL & AWT & ATT & AQL & AWT \\
        \midrule
        LightGPT (Llama3-8b) & \underline{277.38} & \underline{164.60} & \underline{48.68} & \underline{270.14} & \underline{103.77} & \underline{44.10} & \underline{262.67} & \underline{130.65} & \underline{44.98} & \textbf{316.12} & \textbf{61.02} & \textbf{53.84} & \underline{333.95} & \textbf{190.91} & 72.57 \\
        JS-GRPO-h3 (Llama3-8b) & 279.59 & 168.24 & 52.80 & 271.59 & 105.43 & 47.57 & 264.74 & 134.01 & 49.48 & 317.48 & 61.89 & 55.04 & 334.76 & 194.48 & \underline{72.43} \\
        JS-GRPO-h6 (Llama3-8b) & 277.73 & 164.92 & 50.28 & 271.89 & 105.83 & 46.59 & 263.47 & 132.39 & 49.72 & 317.95 & 61.93 & \underline{53.94} & 336.45 & 197.08 & 75.03 \\
        DGLight (Llama3-8b) & \textbf{275.92} & \textbf{162.64} & \textbf{47.17} & \textbf{269.68} & \textbf{103.51} & \textbf{43.78} & \textbf{262.16} & \textbf{129.77} & \textbf{43.53} & \underline{316.59} & \underline{61.08} & 54.43 & \textbf{333.14} & \underline{191.61} & \textbf{70.63} \\
        \bottomrule
    \end{tabular}%
    }
\end{table*}

Table~\ref{tab:tsc_llm_results} shows that \textit{DGLight} is the strongest overall method among the four LLM-based controllers. It achieves the best performance on most datasets and metrics, with Hangzhou-1 as the main exception. Overall, this result shows that the proposed method is superior to the reward-discounting baselines and that critic-guided GRPO improves the quality of the traffic-signal decisions produced by the LLM policy.

When compared with the strongest RL baselines in Table~\ref{tab:tsc_baselines}, \textit{DGLight} remains competitive with the CoLight family and even exceeds them on some metrics, showing that the proposed training procedure improves the model's reasoning ability for traffic situations. The main exception is Hangzhou-1, where the CoLight family performs exceptionally well because those models are trained directly on that dataset, which may partly reflect overfitting to that setting.

Finally, \textit{JS-GRPO-h6} generally performs better than \textit{JS-GRPO-h3}, suggesting that a longer horizon for discounted cumulative reward provides a better estimate of the true long-term reward.

\subsection{Transferability to Unseen Cities (RQ2)}
The transferability results show that \textit{DGLight} retains strong performance on datasets that were not used to fit the DQN critic, namely all three Jinan datasets and Hangzhou-2. More importantly, the magnitude of improvement is relatively larger on these unseen datasets than on the datasets most closely aligned with the critic, especially when compared with both the RL baselines and \textit{LightGPT}. This pattern suggests that the proposed method does not merely adapt to the critic's judgment, but instead improves the model's ability to reason about traffic states in a way that generalizes beyond the data seen by the critic.

\subsection{Interpretability of Model Decisions (RQ3)}
Figure~\ref{fig:dglight_reasoning_example} shows a representative reasoning trace generated by \textit{DGLight} for a single traffic state. The example illustrates how the model first describes the relevant traffic conditions and then maps that assessment to a final signal choice. This makes it possible to inspect whether the stated rationale is consistent with the selected action.

\begin{figure}[ht]
    \centering
    \fbox{%
        \begin{minipage}{0.92\linewidth}
        \small
        \ttfamily
        \raggedright
        \#\#\# Step 1: Analysis

        The traffic congestion is primarily dictated by early queued vehicles, with the most significant impact. Therefore, we should pay the most attention to lanes with long queue lengths.
        \newline
        \newline
        Looking at the current state of the intersection, there are no early queued vehicles in the eastern and western through lanes, northern and southern through lanes, eastern and western left-turn lanes, and northern and southern left-turn lanes.
        \newline
        \newline
        However, there are approaching vehicles in different segments of the allowed lanes of each signal. The eastern and western through lanes have a total of 4 approaching vehicles, with 1 in the eastern and 3 in the western. The northern and southern through lanes have a total of 1 approaching vehicle, all in the northern. The eastern and western left-turn lanes have a total of 1 approaching vehicle, all in the western. The northern and southern left-turn lanes have no approaching vehicles.
        \newline
        \newline
        Considering the neighbor incoming totals, we see that the eastern and western through lanes, northern and southern through lanes, eastern and western left-turn lanes, and northern and southern left-turn lanes all have 3 known totals of neighbor incoming vehicles, with 2/2 available.
        \newline
        \newline
        Given these conditions, the most effective traffic signal that will most significantly improve the traffic condition during the next phase would be the one that relieves the lanes with the most approaching vehicles, as they are likely to reach the intersection soon.
        \newline
        \newline
        \#\#\# Step 2: Answer

        Based on the analysis, the signal that will most significantly improve the traffic condition during the next phase is the one that relieves the eastern and western through lanes, as it has the most approaching vehicles.
        \newline
        \newline
        <signal>ETWT</signal>
        \end{minipage}%
    }
    \caption{Representative \textit{DGLight} reasoning trace for a single traffic state.}
    \label{fig:dglight_reasoning_example}
\end{figure}

\subsection{Ablation Study (RQ4)}
This ablation isolates two sources of improvement: traffic-specific LLM adaptation and critic-guided reinforcement learning. The untuned \textit{Llama3-8b} measures the capability of the base model without TSC adaptation, \textit{LightGPT} measures the gain from prior TSC-specific training, and \textit{DGLight} measures the additional benefit of the proposed critic-guided GRPO procedure.

\begin{table*}[t]
    \centering
    \caption{Ablation study comparing the untuned base model, \textit{LightGPT}, and successive \textit{DGLight} checkpoints. Lower values are better. Best and second-best values are highlighted in bold and underlined, respectively.}
    \label{tab:ablation}
    \resizebox{\linewidth}{!}{%
    \begin{tabular}{lccccccccccccccc}
        \toprule
        & \multicolumn{9}{c}{Jinan} & \multicolumn{6}{c}{Hangzhou} \\
        \cmidrule(lr){2-10} \cmidrule(lr){11-16}
        Models & \multicolumn{3}{c}{1} & \multicolumn{3}{c}{2} & \multicolumn{3}{c}{3} & \multicolumn{3}{c}{1} & \multicolumn{3}{c}{2} \\
        \cmidrule(lr){2-4} \cmidrule(lr){5-7} \cmidrule(lr){8-10} \cmidrule(lr){11-13} \cmidrule(lr){14-16}
        & ATT & AQL & AWT & ATT & AQL & AWT & ATT & AQL & AWT & ATT & AQL & AWT & ATT & AQL & AWT \\
        \midrule
        Llama3-8b & 1064.16 & 1346.71 & 804.81 & 1173.01 & 1068.65 & 811.28 & 1077.41 & 1176.40 & 823.16 & 1090.97 & 610.32 & 728.08 & 728.00 & 711.15 & 575.44 \\
        LightGPT (Llama3-8b) & 277.38 & 164.60 & 48.68 & 270.14 & 103.77 & 44.10 & \underline{262.67} & \underline{130.65} & 44.98 & \underline{316.12} & \textbf{61.02} & \textbf{53.84} & \underline{333.95} & \textbf{190.91} & 72.57 \\
        DGLight-iter0 (LightGPT) & \textbf{275.62} & \textbf{162.38} & \textbf{47.16} & \textbf{268.61} & \textbf{102.25} & \textbf{43.31} & \underline{262.37} & 130.72 & \underline{44.40} & \textbf{316.09} & 61.26 & 58.47 & 334.40 & 191.65 & \textbf{69.02} \\
        DGLight-iter1 (LightGPT) & \underline{275.92} & \underline{162.64} & \underline{47.17} & \underline{269.68} & \underline{103.51} & \underline{43.78} & \textbf{262.16} & \textbf{129.77} & \textbf{43.53} & 316.59 & \underline{61.08} & \underline{54.43} & \textbf{333.14} & \underline{191.61} & \underline{70.63} \\
        \bottomrule
    \end{tabular}%
    }
\end{table*}

Here, \textit{DGLight-iter0} denotes the model after one round of rollout collection followed by one round of GRPO optimization, while \textit{DGLight-iter1} denotes the model after two such rounds. The ablation shows a clear overall trend: moving from the untuned \textit{Llama3-8b} model to \textit{LightGPT} yields the largest improvement, confirming the value of traffic-specific adaptation, and critic-guided GRPO then provides additional gains beyond the inherited \textit{LightGPT} checkpoint. The comparison between \textit{DGLight-iter0} and \textit{DGLight-iter1} suggests that the first round already captures most of the improvement, while the second round produces smaller and more mixed changes that mainly refine an already strong policy.

\section{Discussion}
In the interim report, I proposed using \textit{Group-in-Group Policy Optimization (GiGPO)} for the TSC task, by grouping together near-identical traffic intersection states before computing group-relative advantages. Specifically, the idea was to define a tolerance threshold on lane vehicle counts and treat two states as belonging to the same group when the vehicle numbers on every incoming lane differed by no more than that threshold. The motivation was that larger groups of comparable states might improve the stability of advantage estimation during optimization.

In the final system, however, this idea was not carried out. The first reason is practical: in the benchmark traffic datasets used in this study, the number of vehicles on each lane is usually small, often fewer than ten. Under this scale, it is difficult to choose a tolerance threshold that is both meaningful and robust.

The second reason is methodological and is ultimately more important. The implemented rollout collection method does not store trajectory-specific cumulative rewards in the way GiGPO is designed to exploit. Instead, for each observed traffic state, the pipeline records the critic-estimated action values for all admissible signal phases. During the GRPO optimization step, groups can therefore be generated directly by sampling multiple completions for the same prompt and comparing the actions they propose under the stored action-value vector. In other words, the GRPO groups in the final design can be chosen largely according to compute-capacity considerations during optimization rather than by retroactively matching similar states across trajectories.

In summary, GiGPO enlarges GRPO comparison groups so that the resulting advantage estimates are more informative, especially when rewards are sparse and only observed at the trajectory level. That motivation is substantially weaker in \textit{DGLight}, because the training signal is already available in a dense per-state form through the critic's action values. For this reason, adding GiGPO would have increased implementation complexity without bringing significant improvement.
\section{Conclusion and Future Work}

\subsection{Summary of Contributions}
This dissertation presented \textit{DGLight}, a critic-guided reinforcement-learning framework for adapting a large language model to traffic signal control. The study combined a reimplemented \textit{CoLight}-based DQN critic with a GRPO-based policy optimization procedure so that a pretrained LLM could be trained using dense traffic-aware action values rather than only short-horizon environment rewards. The resulting pipeline converts structured intersection states into textual prompts, generates interpretable reasoning traces together with signal decisions, and reuses stored per-action value vectors to support efficient offline policy optimization.

Empirically, the study showed that \textit{DGLight} is the strongest overall method among the compared LLM-based controllers on the selected CityFlow benchmarks, while remaining competitive with strong RL baselines. The results also suggest that critic-guided GRPO improves transferability beyond the city used to fit the critic and preserves the interpretability benefit of language-based traffic control. Taken together, these findings show that value-based traffic knowledge can be transferred into an LLM policy without discarding human-readable reasoning.

\subsection{Future Work}
\subsubsection{Token-Level Reward Assignment}
The current training pipeline assigns a reward at the completion level, which means that all generated tokens are optimized under the same response-level signal. In practice, however, some tokens mainly serve formatting purposes, such as section headers, boilerplate phrases, or the fixed decision tag format. These tokens should not necessarily receive the same credit assignment as the tokens that express traffic analysis or the final signal choice. A promising extension is therefore to introduce token-level reward assignment or token masking, so that structurally required formatting tokens are excluded from optimization or given reduced weight, while informative reasoning tokens and action-selection tokens receive stronger supervision. This may improve training efficiency by aligning the optimization signal more closely with the parts of the output that actually matter for decision quality.

\subsubsection{MLLM-Style Training with a Traffic Situation Encoder}
Another promising direction is to move beyond purely text-serialized traffic states and train an MLLM-style architecture with a dedicated traffic situation encoder. Instead of describing lane counts, queue lengths, and neighboring conditions only through hand-crafted text, a learned encoder could first transform the structured traffic state into a dense embedding that captures the spatial and numerical relationships in the intersection. The language model could then condition on this embedding to generate the same text-based reasoning and final signal decision as in the current framework. Such a design may reduce prompt engineering overhead, shorten the effective input length, and preserve more of the original traffic structure, while still retaining the key advantage of interpretable natural-language outputs.

\printbibliography

@misc{schulman2017proximalpolicyoptimizationalgorithms,
      title={Proximal Policy Optimization Algorithms}, 
      author={John Schulman and Filip Wolski and Prafulla Dhariwal and Alec Radford and Oleg Klimov},
      year={2017},
      eprint={1707.06347},
      archivePrefix={arXiv},
      primaryClass={cs.LG},
      url={https://arxiv.org/abs/1707.06347}, 
}

@article{webster,
      author = {Yin, Biao and Dridi, Mahjoub and Moudni, Abdellah},
      year = {2015},
      month = {12},
      pages = {},
      title = {Traffic network micro-simulation model and control algorithm based on approximate dynamic programming},
      volume = {10},
      journal = {IET Intelligent Transport Systems},
      doi = {10.1049/iet-its.2015.0108}
}

@article{MERCADER2020275,
      title = {Max-pressure traffic controller based on travel times: An experimental analysis},
      journal = {Transportation Research Part C: Emerging Technologies},
      volume = {110},
      pages = {275-290},
      year = {2020},
      issn = {0968-090X},
      doi = {https://doi.org/10.1016/j.trc.2019.10.002},
      url = {https://www.sciencedirect.com/science/article/pii/S0968090X19307442},
      author = {Pedro Mercader and Wasim Uwayid and Jack Haddad}
}

@inproceedings{10.1145/3357384.3357902,
      author = {Wei, Hua and Xu, Nan and Zhang, Huichu and Zheng, Guanjie and Zang, Xinshi and Chen, Chacha and Zhang, Weinan and Zhu, Yanmin and Xu, Kai and Li, Zhenhui},
      title = {CoLight: Learning Network-level Cooperation for Traffic Signal Control},
      year = {2019},
      isbn = {9781450369763},
      publisher = {Association for Computing Machinery},
      address = {New York, NY, USA},
      url = {https://doi.org/10.1145/3357384.3357902},
      doi = {10.1145/3357384.3357902},
      booktitle = {Proceedings of the 28th ACM International Conference on Information and Knowledge Management},
      pages = {1913–1922},
      numpages = {10},
      location = {Beijing, China},
      series = {CIKM '19}
}

@misc{ruan2024coslightcooptimizingcollaboratorselection,
      title={CoSLight: Co-optimizing Collaborator Selection and Decision-making to Enhance Traffic Signal Control}, 
      author={Jingqing Ruan and Ziyue Li and Hua Wei and Haoyuan Jiang and Jiaming Lu and Xuantang Xiong and Hangyu Mao and Rui Zhao},
      year={2024},
      eprint={2405.17152},
      archivePrefix={arXiv},
      primaryClass={cs.MA},
      url={https://arxiv.org/abs/2405.17152}, 
}

@article{Chen_Wei_Xu_Zheng_Yang_Xiong_Xu_Li_2020,
  title        = {Toward A Thousand Lights: Decentralized Deep Reinforcement Learning for Large-Scale Traffic Signal Control},
  volume       = {34},
  url          = {https://ojs.aaai.org/index.php/AAAI/article/view/5744},
  doi          = {10.1609/aaai.v34i04.5744},
  abstractnote = {&lt;p&gt;Traffic congestion plagues cities around the world. Recent years have witnessed an unprecedented trend in applying reinforcement learning for traffic signal control. However, the primary challenge is to control and coordinate traffic lights in large-scale urban networks. No one has ever tested RL models on a network of more than a thousand traffic lights. In this paper, we tackle the problem of multi-intersection traffic signal control, especially for large-scale networks, based on RL techniques and transportation theories. This problem is quite difficult because there are challenges such as scalability, signal coordination, data feasibility, etc. To address these challenges, we (1) design our RL agents utilizing ‘pressure’ concept to achieve signal coordination in region-level; (2) show that implicit coordination could be achieved by individual control agents with well-crafted reward design thus reducing the dimensionality; and (3) conduct extensive experiments on multiple scenarios, including a real-world scenario with 2510 traffic lights in Manhattan, New York City &lt;sup&gt;1&lt;/sup&gt; &lt;sup&gt;2&lt;/sup&gt;.&lt;/p&gt;},
  number       = {04},
  journal      = {Proceedings of the AAAI Conference on Artificial Intelligence},
  author       = {Chen, Chacha and Wei, Hua and Xu, Nan and Zheng, Guanjie and Yang, Ming and Xiong, Yuanhao and Xu, Kai and Li, Zhenhui},
  year         = {2020},
  month        = {Apr.},
  pages        = {3414-3421}
}

@inproceedings{pmlr-v162-zhang22ah,
  title     = {Expression might be enough: representing pressure and demand for reinforcement learning based traffic signal control},
  author    = {Zhang, Liang and Wu, Qiang and Shen, Jun and L{\"u}, Linyuan and Du, Bo and Wu, Jianqing},
  booktitle = {Proceedings of the 39th International Conference on Machine Learning},
  series    = {Proceedings of Machine Learning Research},
  volume    = {162},
  pages     = {26645--26654},
  year      = {2022},
  month     = {17--23 Jul},
  publisher = {PMLR},
  url       = {https://proceedings.mlr.press/v162/zhang22ah.html}
}

@misc{tang2019cityflowcityscalebenchmarkmultitarget,
      title={CityFlow: A City-Scale Benchmark for Multi-Target Multi-Camera Vehicle Tracking and Re-Identification}, 
      author={Zheng Tang and Milind Naphade and Ming-Yu Liu and Xiaodong Yang and Stan Birchfield and Shuo Wang and Ratnesh Kumar and David Anastasiu and Jenq-Neng Hwang},
      year={2019},
      eprint={1903.09254},
      archivePrefix={arXiv},
      primaryClass={cs.CV},
      url={https://arxiv.org/abs/1903.09254}, 
}

@misc{wang2024llmassistedlightleveraging,
      title={LLM-Assisted Light: Leveraging Large Language Model Capabilities for Human-Mimetic Traffic Signal Control in Complex Urban Environments}, 
      author={Maonan Wang and Aoyu Pang and Yuheng Kan and Man-On Pun and Chung Shue Chen and Bo Huang},
      year={2024},
      eprint={2403.08337},
      archivePrefix={arXiv},
      primaryClass={eess.SY},
      url={https://arxiv.org/abs/2403.08337}, 
}

@misc{pang2024illmtscintegrationreinforcement,
      title={iLLM-TSC: Integration Reinforcement Learning and Large Language Model for Traffic Signal Control Policy Improvement}, 
      author={Aoyu Pang and Maonan Wang and Man-On Pun and Chung Shue Chen and Xi Xiong},
      year={2024},
      eprint={2407.06025},
      archivePrefix={arXiv},
      primaryClass={cs.AI},
      url={https://arxiv.org/abs/2407.06025}, 
}

@misc{zhang2026latslargelanguagemodelassisted,
      title={LATS: Large Language Model Assisted Teacher-Student Framework for Multi-Agent Reinforcement Learning in Traffic Signal Control}, 
      author={Yifeng Zhang and Peizhuo Li and Tingguang Zhou and Mingfeng Fan and Guillaume Sartoretti},
      year={2026},
      eprint={2603.24361},
      archivePrefix={arXiv},
      url={https://arxiv.org/abs/2603.24361}, 
}

@misc{ouyang2022traininglanguagemodels,
      title={Training Language Models to Follow Instructions with Human Feedback}, 
      author={Long Ouyang and Jeff Wu and Xu Jiang and Diogo Almeida and Carroll L. Wainwright and Pamela Mishkin and Chong Zhang and Sandhini Agarwal and Katarina Slama and Alex Ray and others},
      year={2022},
      eprint={2203.02155},
      archivePrefix={arXiv},
      primaryClass={cs.CL},
      url={https://arxiv.org/abs/2203.02155}, 
}

@misc{schulman2015trustregionpolicyoptimization,
      title={Trust Region Policy Optimization}, 
      author={John Schulman and Sergey Levine and Philipp Moritz and Michael I. Jordan and Pieter Abbeel},
      year={2015},
      eprint={1502.05477},
      archivePrefix={arXiv},
      primaryClass={cs.LG},
      url={https://arxiv.org/abs/1502.05477}, 
}

@misc{shao2024deepseekmath,
      title={DeepSeekMath: Pushing the Limits of Mathematical Reasoning in Open Language Models}, 
      author={Zhihong Shao and Peiyi Wang and Qihao Zhu and Runxin Xu and Junxiao Song and Mingchuan Zhang and Y. K. Li and Y. Wu and Daya Guo and Dejian Yang and others},
      year={2024},
      eprint={2402.03300},
      archivePrefix={arXiv},
      primaryClass={cs.CL},
      url={https://arxiv.org/abs/2402.03300}, 
}

@misc{rafailov2023directpreferenceoptimization,
      title={Direct Preference Optimization: Your Language Model is Secretly a Reward Model}, 
      author={Rafael Rafailov and Archit Sharma and Eric Mitchell and Stefano Ermon and Christopher D. Manning and Chelsea Finn},
      year={2023},
      eprint={2305.18290},
      archivePrefix={arXiv},
      primaryClass={cs.LG},
      url={https://arxiv.org/abs/2305.18290}, 
}

@misc{lai2024llmlightlargelanguagemodels,
      title={LLMLight: Large Language Models as Traffic Signal Control Agents}, 
      author={Siqi Lai and Zhao Xu and Weijia Zhang and Hao Liu and Hui Xiong},
      year={2024},
      eprint={2312.16044},
      archivePrefix={arXiv},
      primaryClass={cs.AI},
      url={https://arxiv.org/abs/2312.16044}, 
}

@misc{zou2025trafficr1reinforcedllmsbring,
      title={Traffic-R1: Reinforced LLMs Bring Human-Like Reasoning to Traffic Signal Control Systems}, 
      author={Xingchen Zou and Yuhao Yang and Zheng Chen and Xixuan Hao and Yiqi Chen and Chao Huang and Yuxuan Liang},
      year={2025},
      eprint={2508.02344},
      archivePrefix={arXiv},
      primaryClass={cs.AI},
      url={https://arxiv.org/abs/2508.02344}, 
}

@misc{guo2025deepseekr1,
      title={DeepSeek-R1: Incentivizing Reasoning Capability in LLMs via Reinforcement Learning}, 
      author={Daya Guo and Dejian Yang and Haowei Zhang and Junxiao Song and Ruoyu Zhang and Runxin Xu and Qihao Zhu and Shengqi Li and P. Wang and others},
      year={2025},
      eprint={2501.12948},
      archivePrefix={arXiv},
      primaryClass={cs.CL},
      url={https://arxiv.org/abs/2501.12948}, 
}

\appendix
\section{Appendix}

\subsection{Joint-Scored GRPO Comparison Variant}
\label{appendix:joint_scored_grpo}
This appendix summarizes the \textit{Joint-Scored GRPO} pipeline used as a comparison variant to the proposed \textit{DGLight} method. The purpose of this variant is to test whether indirect short-horizon joint return estimation can provide a better training signal than direct critic guidance. Unlike the main method, it does not rely on a pretrained action-value critic to score each sampled response, but uses the trajectory-aggregated raw environment rewards to compute group advantages.

\paragraph{Snapshot-forked joint candidate evaluation.}
At each decision step, the policy samples $k$ candidate responses for every intersection and parses them into actions. Instead of scoring each sampled action independently, the method aligns candidates across intersections into $k$ joint action vectors (each with actions for all intersections) and evaluates each vector by forking the current simulator snapshot. Each fork starts from the same environment state but executes a different sampled joint action, allowing the method to compare candidate action sets under a common traffic condition.

\paragraph{Short-horizon discounted return.}
For each forked candidate $c$, the simulator is rolled out for a short horizon $H$. After the first jointly sampled action is executed, subsequent actions are resampled from the current policy at later steps. The candidate score is the discounted return
\[
G_t^{(c)} = \sum_{\tau=0}^{H-1} \gamma^\tau r_{t+\tau}^{(c)},
\]
where $\gamma$ is the discount factor and $r_{t+\tau}^{(c)}$ is the congestion-based network reward observed at rollout step $t+\tau$ in candidate fork $c$.

\paragraph{Local, neighborhood, and global reward mixing.}
The per-step reward is computed from a mixed congestion cost that combines local intersection congestion, neighboring congestion, and global network congestion. Let $m_i$ denote the local congestion metric at intersection $i$ (for example, incoming queue length or pressure), let $\bar m_{\mathcal{N}(i)}$ denote the mean metric over the neighbors of $i$, and let $\bar m_{\mathcal{V}}$ denote the mean metric over the whole network. The reward is defined from the negative mixed cost
\[
r_t = - \frac{1}{|\mathcal{V}|} \sum_{i \in \mathcal{V}}
\Bigl(
\alpha m_i
  + \beta \bar m_{\mathcal{N}(i)}
  + (1-\alpha-\beta)\bar m_{\mathcal{V}}
\Bigr),
\]
where $\alpha$ and $\beta$ are hyperparameters that control the relative contribution of local and neighborhood congestion. This construction encourages decisions that reduce both immediate local congestion and broader network imbalance.

\paragraph{Projection back to per-intersection action scores.}
Because GRPO training still expects prompt-level rewards for individual sampled responses, the joint candidate returns must be projected back to action-level scores. For each intersection $i$ and each action $a \in A$, let $\mathcal{C}_{i,a}$ be the set of joint candidates whose parsed action at intersection $i$ equals $a$. The induced action score is defined as the mean return of those candidates:
\[
\tilde Q_{i,t}(a) =
\frac{1}{|\mathcal{C}_{i,a}|}
\sum_{c \in \mathcal{C}_{i,a}} G_t^{(c)},
\]
with a fallback reward used when no sampled candidate proposes action $a$. The resulting vector $\tilde{\mathbf{q}}_{i,t}$ plays the same role in GRPO training as the critic score vector $\mathbf{q}_{i,t}$ in the main method, but it is obtained indirectly through joint rollout evaluation.

\paragraph{Collect-then-train split.}
In implementation, the variant is organized as a collect-then-train pipeline. A rollout stage first saves prompt records, selected actions, and projected action-score vectors obtained from snapshot-forked candidate evaluation. The GRPO stage then reloads these records and fine-tunes the LLM policy using the same prompt-conditioned reward lookup mechanism as in the main method.

\subsection{Sample Prompt}
\label{appendix:sample_prompt}
You are an expert in traffic management.

A traffic light regulates a four-way intersection with northern, southern, eastern, and western approaches, each containing two lanes: one for through traffic and one for left-turns. Each lane is further divided into three segments. Segment 1 is the closest to the intersection. Segment 2 is in the middle. Segment 3 is the farthest. In a lane, there may be early queued vehicles and approaching vehicles traveling in different segments. Early queued vehicles have already arrived at the intersection and await passage permission. Approaching vehicles will arrive at the intersection in the future.

The traffic light has 4 signal phases. Each signal relieves vehicles' flow in a group of two specific lanes.

Available signal phases:

- ETWT: Eastern and western through lanes.

- NTST: Northern and southern through lanes.

- ELWL: Eastern and western left-turn lanes.

- NLSL: Northern and southern left-turn lanes.

Current intersection state:

Signal: ETWT

Relieves: Eastern and western through lanes.

- Early queued: 0 (East), 1 (West), 1 (Total)

- Segment 1: 0 (East), 0 (West), 0 (Total)

- Segment 2: 1 (East), 0 (West), 1 (Total)

- Segment 3: 0 (East), 2 (West), 2 (Total)

- Neighbor incoming totals: 2 (East), 1 (West), 3 (Known total), 2/2 available

Signal: NTST

Relieves: Northern and southern through lanes.

- Early queued: 0 (North), 0 (South), 0 (Total)

- Segment 1: 0 (North), 0 (South), 0 (Total)

- Segment 2: 0 (North), 0 (South), 0 (Total)

- Segment 3: 1 (North), 0 (South), 1 (Total)

- Neighbor incoming totals: NA (North), 3 (South), 3 (Known total), 1/2 available

Signal: ELWL

Relieves: Eastern and western left-turn lanes.

- Early queued: 0 (East), 0 (West), 0 (Total)

- Segment 1: 0 (East), 0 (West), 0 (Total)

- Segment 2: 0 (East), 0 (West), 0 (Total)

- Segment 3: 0 (East), 1 (West), 1 (Total)

- Neighbor incoming totals: 2 (East), 1 (West), 3 (Known total), 2/2 available

Signal: NLSL

Relieves: Northern and southern left-turn lanes.

- Early queued: 0 (North), 0 (South), 0 (Total)

- Segment 1: 0 (North), 0 (South), 0 (Total)

- Segment 2: 0 (North), 0 (South), 0 (Total)

- Segment 3: 0 (North), 0 (South), 0 (Total)

- Neighbor incoming totals: NA (North), 3 (South), 3 (Known total), 1/2 available

The state description above lists:

- The group of lanes relieved under each traffic light phase.

- The number of early queued vehicles in the allowed lanes of each signal.

- The number of approaching vehicles in different segments of the allowed lanes of each signal.

- Neighbor incoming totals from adjacent intersections for each phase.

- `NA` means that adjacent side has a virtual/missing neighbor and is excluded from `Known total`.

Question:

Which is the most effective traffic signal that will most significantly improve the traffic condition during the next phase?

Note:

- Traffic congestion is primarily dictated by early queued vehicles, with the most significant impact.

- You must pay the most attention to lanes with long queue lengths.

- It is not urgent to consider vehicles in distant segments, since they are unlikely to reach the intersection soon.

Requirements:

- Think step by step.

- You can only choose one of the signals listed above.

- Step 1: Provide a brief analysis identifying the optimal traffic signal.

- Step 2: After finishing the analysis, answer with your chosen signal.

- Include exactly one final decision tag in this format: <signal>PHASE</signal>, where PHASE is one of: ETWT, NTST, ELWL, NLSL.

\end{document}